\documentclass[11pt,a4paper]{article}
\usepackage[hyperref]{acl2017}
\usepackage{times}
\usepackage{latexsym}

\usepackage{url}
\usepackage{verbatim}
\usepackage{bm}
\usepackage{amsmath,amssymb}
\usepackage{multirow}
\usepackage{graphicx}
\usepackage{tabularx}
\usepackage{booktabs}
\usepackage{color}

\aclfinalcopy % Uncomment this line for the final submission
\def\aclpaperid{3101} %  Enter the acl Paper ID here

\title{Multi-Modal Similarity Metric Learning for Answer Selection}
\author{Lingxun Meng$^1$\quad Yan Li$^2$\\
$^1$Shenma Search, Alibaba Mobile Group,\\
Alibaba-Inc, Beijing, 100083, China.\\
{\tt lingxun.mlx@alibaba-inc.com}\\
$^2$Key Lab of Intelligent Information Processing of Chinese Academy of Sciences (CAS), \\
Institute of Computing Technology, CAS, Beijing, 100190, China. \\
{\tt yan.li@vipl.ict.ac.cn}\\}

\date{}

\begin{document}
\maketitle
\begin{abstract}
Recent works using artificial neural networks based on distributed word representation greatly boost performance on answer selection problem. Nevertheless, most of the previous works used deep learning methods (like LSTM-RNN, CNN, etc.) mainly to capture semantic representation of each sentence separately, ignoring the interdependence between each other on lexical level. In this paper, we constitutes a deep convolutional network directly on pairwise token matching, multi-modal similarity metric learning is then adopted to enrich the lexical modality matching. The proposed model demonstrates its performance by surpassing previous state-of-the-art systems on the answer selection benchmark, i.e., TREC-QA dataset, in both MAP and MRR metrics.
\end{abstract}

\section{Introduction}

Inspired by the achievements of convolutional networks (a.k.a, ConvNets) in the field of computer vision, more and more researchers constitute ConvNets for kinds of natural language processing tasks, e.g., text classification \cite{kim2014convolutional}, text regression \cite{bitvai2015non}, short text pair re-ranking \cite{severyn2015learning}, and semantic matching \cite{hu2014convolutional}. \\
\indent For the answer selection task, i.e., given a question and a set of candidate sentences, choose the correct sentence that contains the exact answer and sufficiently support the answer choices. Most of the previous methods constitute Siamese-like deep architectures (like LSTM-RNN, CNN, etc.) to learn the semantic representation for each sentence, and then use cosine similarity or weight matrix to compute the similarity of the pairwise representations \cite{wang2015long}. At the same time, these works mostly adopted shallow architectures for sentence modeling, since deeper nets did not bring better performance. On the contrary, we firmly convinced that one can benefit much more from deep learning strategy.\\
\indent Following the success of RNN-based attentive mechanism designed for machine translation task \cite{bahdanau2014neural}, recently some works attempted two-way attention mechanism for sentence pair matching problems \cite{tan2015lstm,santos2016attentive,yin2015abcnn}. Such soft attention mechanism proves the effectiveness of the interaction between sentence pairs from lexical level to semantic level, yet aggravates much more computations and model complexity.\\
\begin{table}%[!htb]
\setlength{\abovecaptionskip}{0.3cm}
\setlength{\belowcaptionskip}{-0.55cm}
\begin{tabularx}{7.7cm}{lX}
%\hline
Q: & When did Amtrak {\bf{\em \color{red} \underline {begin}}} operations?\\%Where do you put the {\bf  {\em \underline {apple}}}?\\
A: & Amtrak has not turned a profit since it was {\bf{\em \color{red} \underline {founded}}} in 1971.\\%{\bf {\em \underline{ Apple}}} is an American company headquartered in Cupertino, California.\\
%A2: & {\bf  {\em \underline {Apples}}} are stored in the cupboard.\\
%\hline
\end{tabularx}
\caption{An example of QA-pair in TREC-QA.}
\label{tab:example}
\end{table}
\begin{figure*}
\setlength{\abovecaptionskip}{0.0cm}
\setlength{\belowcaptionskip}{-0.35cm}
\centering
\includegraphics[width=16.0cm]{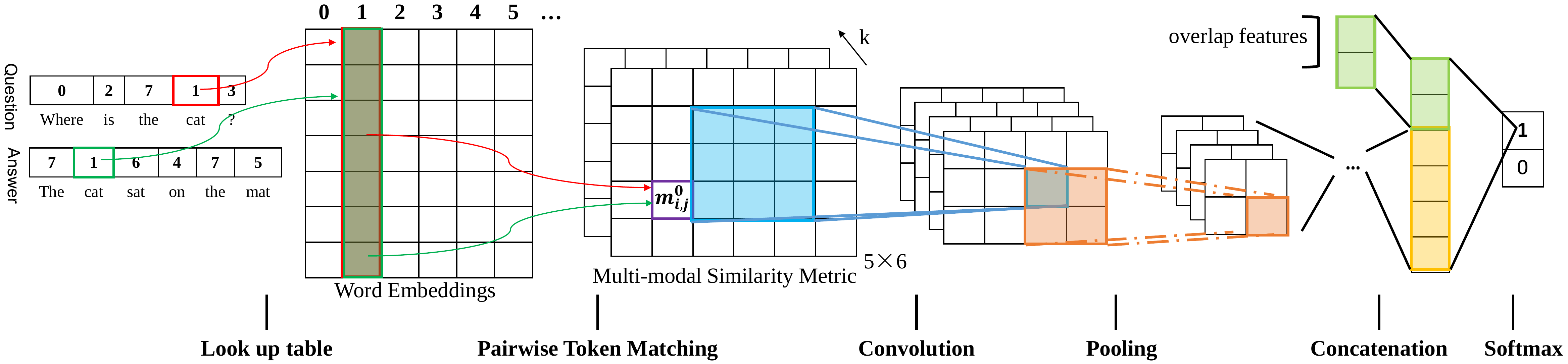}
\label{fig:thewholeframework}
\caption{Our M$^2$S-Net for sentence pair matching.}
\end{figure*}
\indent All the previous works mentioned above motivate us to construct a network based on pairwise token matching for exhaustive matching learning. However, a vital issue of this constitution is the word similarity measurement.Take Q and A in Table \ref{tab:example} for example, distinguishing the similarity between "begin" with "found: set up" from that between "begin" with "found: discovered" makes a lot of sense. To solve this, we constitute a deep convolutional neural network based on pairwise token matching measured with multi-modal similarity metric learning, named by us M$^2$S-Net, where the learnable multi-modal similarity metric provides a comprehensive and multi-granularity measurement. Experimental results on the benchmark dataset of answer selection task indicate that the proposed model can greatly benefit from deep network structure as well as multi-modal similarity metric learning, and also demonstrate that the proposed M$^2$S-Net outperforms a variety of strong baselines and achieve state-of-the-art.
\section{M$^2$S-Net}
In this paper, we propose a novel learning framework for sentence pair matching, where the pairwise token similarity matrix is computed firstly, and then a deep convolutional network is constructed to learn matching representation exhaustively, finally concatenate the learned pairwise matching representation and additional simple word-level overlap feature to feed into a {\em pointwise} rank loss for end-to-end fine-tuning (please see Fig. \ref{fig:thewholeframework} for better understanding).
\subsection{Multi-Modal Similarity Metric}
As a fundamental component in M$^2$S-Net, it is of vital importance to design an appropriate similarity measurement $f_{match}$ for pairwise token matching. Given a sentence pair $S_1 = \{{\bm w}_1^{i}, i \in [0, l_{1}-1]\}$ and $S_2 = \{{\bm w}_2^{j},  j \in [0, l_{2}-1]\}$, where $l_{1}, l_{2}$ are the word count of sentence $S_1, S_2$, respectively, ${\bm w}_1^i, {\bm w}_2^j$ are from $d$-dimensional word embeddings $W$ which are pre-trained under vocabulary $V$, The similarity matrix $M \in \mathbb{R}^{k \times l_{1}\times l_{2}}$ is formulated as follows: 
\begin{comment}
\begin{align}
\begin{small}
m_{i,j} = \left\{
\begin{array}{l@{}c@{}l}
1/({1+||{\bm w}_i^1-{\bm w}_j^2||})  &   &   ({\textrm {Euc.}}) \\
({{{\bm w}_i^1}^T{\bm w}_j^2})/({||{\bm w}_i^1||\cdot ||{\bm w}_j^2||}) & & ({\textrm {Cos.}}) \label {eq:1}\\
{{\bm w}_i^1}^T{U}{\bm w}_j^2+b_{i,j}& & ({\textrm {Metric.}})\\
\end{array} \right.
\end{small}
\end{align}
\end{comment}
\begin{equation}
M = [m^k_{i,j}] = [{{\bm w}_i^1}^T{U^k}{\bm w}_j^2+b^k_{i,j}]
\end{equation}
where k represents the number of modality that can be tuned, ${U^k} \in \mathbb {R}^{d\times d}$ represents the matrix of the learnable similarity metric, and the corresponding bias term ${B}^k \in \mathbb {R}^{l_{1}\times l_{2}}$. Since the dimension of metric $U^k$ increases exponentially with word embedding dimension, some regularization methods were proposed \cite{shalit2010online, cao2013similarity} to limit model complexity for preventing overfitting. Frobenius norm is adopted here for simplification. For better comparison, we also design cosine and euclidean similarity, which are formulated as follows:
\begin{align}
\begin{small}
m_{i,j} = \left\{
\begin{array}{l@{}c@{}r}
1/({1+||{\bm w}_i^1-{\bm w}_j^2||})  &   &   ({\textrm {Euc.}}) \\
({{{\bm w}_i^1}^T{\bm w}_j^2})/({||{\bm w}_i^1||\cdot ||{\bm w}_j^2||}) &  & ({\textrm {Cos.}}) \label {eq:1}\\
\end{array} \right.
\end{small}
\end{align}
\subsection{Convolution and Pooling}
\label{sect:convandpooling}
The convolution layer in this work consists of a filter bank $ F\in \mathbb {R}^{n \times c \times h \times w}$, along with filter biases $\bm b \in \mathbb {R}^n$, where $n$, $w$ and $h$ refer to the number, width and height of filters respectively, and $c$ denotes the channels of data from the lower layer. More specifically, for the first convolution layer, $c$ equals to the multi-modal parameter $k$, which means convolving across all the similarity modalities to learn the pattern. Given the output $L^{t-1} \in \mathbb {R}^{c\times l_h \times l_w}$ ($L^0$ represents similarity matrix $M$) from the lower layer, the output of the convolution with filter bank $F$ is computed as follows:
\begin{equation}
\begin{small}
\begin{aligned}
 L^{t} &= {\tanh}(F * L^{t-1} + {\bm b}) \\
 &={\tanh}([{\bm f}_i^T{\bm l}^{t-1}_{c\times(j-h+1:j)\times(l-w+1:l)} + b_i])\\
\end{aligned}
\label {eq:conv}
\end{small}
\end{equation}
where * is marked as the convolutional operation, $i$ indexes the number of filters, $j$ and $l$ indicates the sliding operations for dot production along the axis of width and height with one step size. \\
\indent Typically, there exist two types of convolution: {\em wide} and {\em narrow}. Even though previous works \cite{KalchbrennerGB14} have pointed out that using {\em wide} type of convolution got better performance, we use the {\em narrow} type for convenience. Finally, we get the output of layer $t$ as $L^{t} \in \mathbb {R}^{n \times (l_h-h+1) \times (l_w-w+1)}$.\\
\indent The outputs from the convolutional layer (passed through the activation function) are then fed into the pooling layer, whose goal is to aggregate the information and reduce the representation. Technically, there exist two types of pooling strategy, i.e., {\em average} pooling and {\em max} pooling, and both pooling methods are widely used. However, {\em max} pooling can lead to strong over-fitting on the training data and, hence, poor generalization on the test data, as shown in \cite{Zeiler_stochpooling}. For stability and reproductivity, we adopt the {\em average} pooling strategy in our work.
\subsection{{\em Pointwise} Learning to Rank with Metric Regularization}
\label{ssec:pwrl}
We adopt simple {\em pointwise} method to model our answer selection task, though {\em pairwise} and {\em listwise} approaches claim to yield better performance. The cross-entropy cost deployed here to discriminantly train our framework as follows:
%\begin{equation}
\begin{multline}
\setlength{\abovedisplayskip}{1pt}
\setlength{\belowdisplayskip}{1pt}
%loss({\bm p}, {\bm y};{\bm \theta})=
C_{\theta}={-\frac {1}{N}\sum_{i=1}^N{[y_i\log p_i+(1-y_i)\log(1-p_i)]}} \\
{\hspace{0mm}}+ \frac{\lambda}{2}{ \Arrowvert U \Arrowvert^2_F}
\label{eq:4}
\end{multline}
%\end{equation}
where $p_i$ is the output probability of $i^{th}$ sample through our networks, $y_i$ is the corresponding ground truth, and $\theta$ contains all the parameters optimized by the network, i.e., $\theta=\{W;U;B;[F];[{\bm b}]\}$. Frobenius norm is used to regularize the parameter $U$ of the metrics to prevent over-fitting.\\% $\lambda$ is set to be $5e^{-4}$.
\indent We use Stochastic Gradient Descent (SGD) to optimize our network, and  AdaDelta \cite{zeiler2012adadelta} is used to automatically adapt the learning rate during the training procedure. For higher performance, hyper-parameter selection is conducted on the development set, and Batch Normalization ({\em BN}) layer \cite{ioffe2015batch} after each convolution layer is also added to speed up the network optimization. In addition, dropout is applied after the first hidden layer for regularization, and early stopping is used to prevent over-fitting with a patience of 5 epochs.
\begin{table}
\setlength{\abovecaptionskip}{0.3cm}
\setlength{\belowcaptionskip}{-0.35cm}
\centering
\small
\begin{tabular}{lrrrl}
\toprule[1.5pt]
Set & \#Question &  \#QApairs &  \%Correct &  Judge\\
\midrule[0.8pt]
%\multirow{3}{*}{{TrecQA}}
Train-All & {1,229} & {53,417} & {12.0\%} & {auto} \\
Train & {94} & {4,718} & {7.4\%} & man \\
Dev & {65} & {1,117} & {18.4\%} & man \\
Test & {68} & {1,442} & {17.2\%} & man\\
\bottomrule[1.5pt]
\end{tabular}
\caption{Statistics of the answer sentence selection dataset. Judge denotes whether correctness was determined automatically (auto) or by human annotators (man).}
\label{tab:datasetsinfo}
\end{table}
\section{Experiments}
\label{sec:exp}
\subsection{Dataset}
\label{ssec:dataset}
In this section, we use TREC-QA dataset to evaluate the proposed model, which appears to be one of the most widely used benchmarks for answer sentence selection. This dataset was created by \cite{wang2007jeopardy} based on Text REtrieval Conference (TREC) QA track (8-13) data\footnote{\url{http://cs.stanford.edu/people/mengqiu/data/qg-emnlp07-data.tgz}}. Candidate answers were automatically retrieved for each factoid question. Two sets of data are provided for training, one is small training set containing 94 questions collected through manual judgement, and the other is full training set, i.e., Train-All, which contains 1,229 questions from the entire TREC 8-12 collection with automatically labeled ground truth by matching answer keys' regular expressions\footnote {\url{http://cs.jhu.edu/~xuchen/packages/jacana-qa-naacl2013-data-results.tar.bz2}}. Table \ref{tab:datasetsinfo} summarizes the answer selection dataset in details. In the following experiments, we use the full training set due to its relatively large scale, even though there exists noisy labels caused by automatically pattern matching.\\
\indent The original development and test datasets have 82 and 100 questions, respectively. Following \cite{wang2015long,santos2016attentive,tan2015lstm}, all questions with only positive or negative answers are removed. Finally, we have 65 development questions with 1,117 question-answer pairs, and 68 test questions with 1,442 question-answer pairs.
\subsection{Token Representation}
\label{sec:tokens}
We use a pre-trained 50-dimensional word embeddings\footnote{\url{http://nlp.stanford.edu/data/glove.6B.zip}} \cite{pennington2014glove} as our initial word look-up table. These word embeddings are trained on Wikipedia data and the fifth English Gigawords with totally 6 Billion tokens. Need to be mentioned here, trading off between model complexity and performance, we do not use the 300-dimensional embeddings, which are trained on much more data and more widely adopted by previous works \cite{santos2016attentive,tan2015lstm}. 
\begin{figure}
\setlength{\abovecaptionskip}{0.1cm}
\setlength{\belowcaptionskip}{-0.35cm}
\centering
\includegraphics[width=8.0cm]{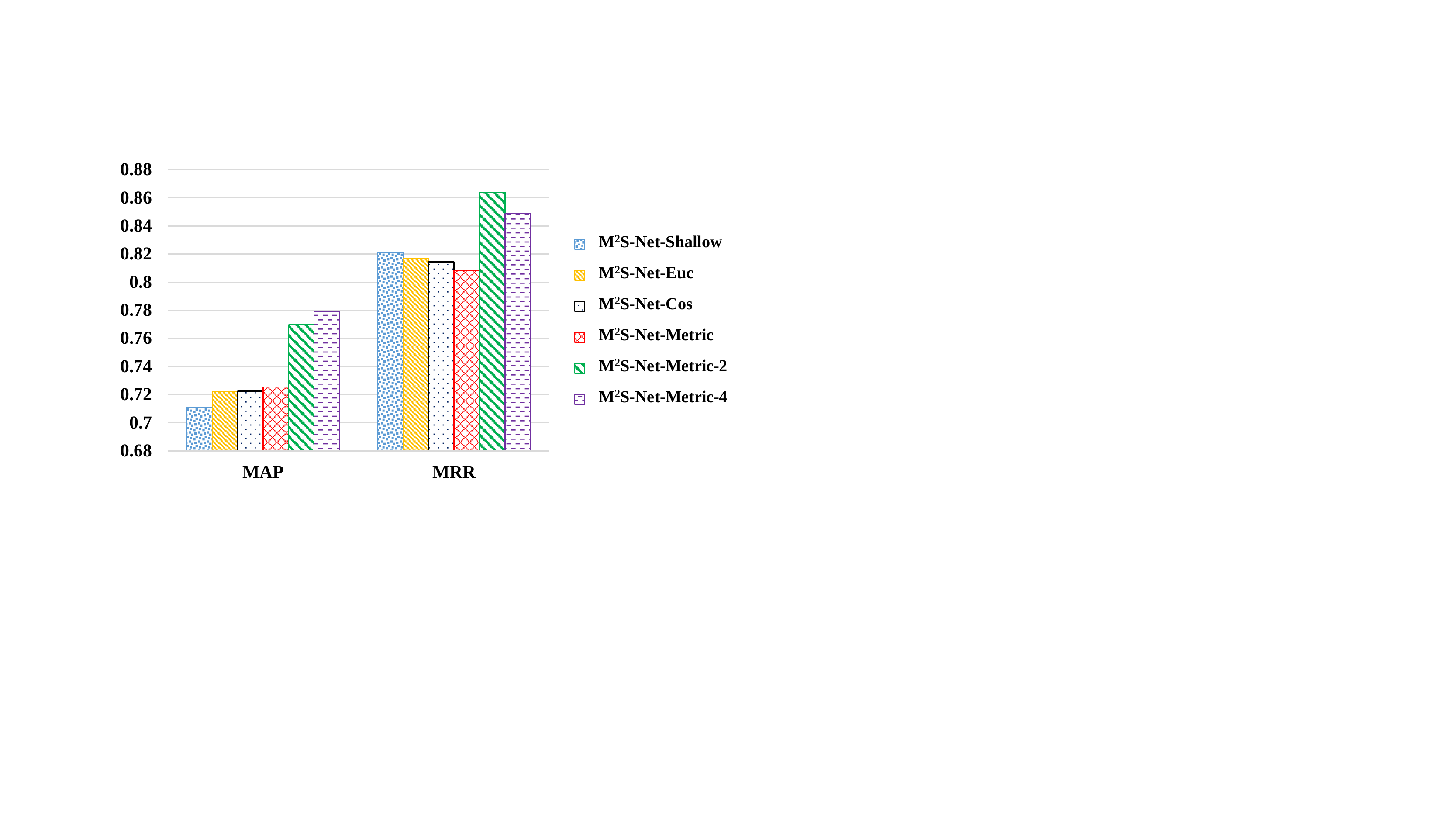}
\caption{Comparison of $M^2$S-Nets with different measurements and network structure.}
\label{fig:comp1}
\end{figure}
\subsection{Experimental Setting}
\label{sec:setting}
Following previous works, we also use the two metrics to evaluate the proposed model, i.e., Mean Average Precision (MAP) and Mean Reciprocal Rank (MRR). The official {\em trec\_eval} scorer tool\footnote{\url{http://trec.nist.gov/trec_eval/}} is used to compute the above metrics.\\
\indent The simplest word overlap features between each question-answer pair are computed, and we concatenate them with our learned matching representation for the final rank learning. This feature vector contains only two features, i.e., word overlap and IDF-weighted word overlap.\\
\indent Experiments of our M$^2$S-Net on three pre-defined similarity measurements are denoted as M$^2$S-Net-Euc, M$^2$S-Net-Cos, and M$^2$S-Net-Metric respectively. All of these models share the same network configuration. To demonstrate the fact that the proposed network can benefit more from deep structure, we compare M$^2$S-Net-Metric with a one-convolutional layer network, namely M$^2$S-Net-Shallow (We found that much deeper construction might bring in randomness which harms the reproductivity of the performance, so we use two-convolutional layer for strict experimental comparison). Furthermore, we list results of M$^2$S-Net-Metric with $k = 1, 2, 4$, respectively denoted as M$^2$S-Net-Metric, M$^2$S-Net-Metric-2 and M$^2$S-Net-Metric-4, to verify the effectiveness of the proposed multi-modal similarity metric. All the networks mentioned here are implemented using Caffe \cite{jia2014caffe} and the code is open now\footnote{\url{https://github.com/lxmeng/mms_answer_selection}}.
\begin{comment}
\begin{table}
\setlength{\abovecaptionskip}{0.3cm}
\setlength{\belowcaptionskip}{-0.35cm}
\centering 
\begin{tabular}{lcc}
\toprule[1.5pt]
Model & MAP & MRR\\  
\midrule[0.6pt]
M$^2$S-Net-Shallow & .7111 & .8210\\
\midrule[1pt]  
M$^2$S-Net-Euc & .7220 & .8172\\ 
M$^2$S-Net-Cos & .7226 & .8144\\
M$^2$S-Net-Metric & .7256 & .8082\\
\midrule[0.6pt]
M$^2$S-Net-Metric-2 & .7698 & {\bf .8640}\\
M$^2$S-Net-Metric-4 & {\bf .7793} &  .8487\\
\bottomrule[1.5pt] 
\end{tabular}
\caption{Results of M$^2$S-Nets on the answer sentence selection dataset.}
\label{tab:rst1}
\end{table}
\end{comment}
\section{Results and Discussion}
\label{sec:rst}
We are motivated to use multi-modal similarity metric to solve polysemy of words, and construct thorough matching  network between sentence pairs for end-to-end question answering modeling. From Fig. \ref{fig:comp1}, we can see that one-modality metric is slightly better than euclidean and cosine similarity measurement. Increasing the number of modality of measurement greatly boost the performance by 7\%. The comparison between shallow and deep network structure indicate that the proposed M$^2$S-Net benefits much from deep construction. The rank of answer in Table \ref{tab:example} is promoted from top 35 and 26 by using euclidean and cosine similarity measuremen to top 3 by using ours.\\
\indent For comprehensive comparison, we also list the results of prior state-of-the-art methods in literature on this task in Table \ref{tab:rst2}. It can be seen that the proposed method outperforms the most recently published attention-based methods by 1\% in both MAP and MRR metrics.\\
\indent The proposed method could be further improved by upgrading the regularization term to limit the rank of metric, which had been proved by \cite{law2014fantope, cao2013similarity}. Besides, combining the dissimilarity modeled by distance metric learning with similarity mentioned here would be our future work.
\begin{table}
\setlength{\abovecaptionskip}{0.3cm}
\setlength{\belowcaptionskip}{-0.35cm}
\centering 
\begin{tabular}{lcc}
\toprule[1.5pt]
 Reference &  MAP & MRR\\ 
\midrule[1pt] 
\newcite{wang2007jeopardy} & .6029 & .6852 \\ 
\newcite{heilman2010tree} & .6091 & .6917\\
\newcite{wang2010probabilistic} & .5951 & .6951\\
\newcite{yao2013answer} & .6307 & .7477\\
\newcite{yih2013question} & .7092 & .7700\\
\newcite{yu2014deep} & .7113 & .7846\\
\newcite{wang2015long} & .7134 & .7913\\
\newcite{tan2015lstm} & .7106 & .7998\\
\newcite{severyn2015learning} & .7459 &.8078\\
\newcite{santos2016attentive} & .7530 & .8511\\
\newcite{wang2016sentence} & .7714 & .8447\\
\midrule[0.6pt]
M$^2$S-Net-Metric-2 & .7698 & {\bf .8640}\\
M$^2$S-Net-Metric-4 & {\bf .7793} &  .8487\\
\bottomrule[1.5pt] 
\end{tabular}
\caption{Results of our models and other methods from the literature.}
\label{tab:rst2}
\end{table}
\section {Conclusion}
\label{sect:ccls}
A novel end-to-end learning framework (M$^2$S-Net) is proposed for answer sentence selection task. Interdependence between sentence pair at lexical level is explored much more by constituting deep convolutional neural network directly on pairwise token matching. To enrich the lexical modality measurement, we adopt multi-modal similarity metric learning. The proposed architecture is proved effective, and surpasses previous state-of-the-art systems on the answer selection benchmark, i.e., TREC-QA dataset, in both MAP and MRR metrics.

% include your own bib file like this:
%\bibliographystyle{acl}
%\bibliography{acl2017}
\bibliography{acl2017}
\bibliographystyle{acl_natbib}

\end{document}